# Bias Amplification in Artificial Intelligence Systems

Kirsten Lloyd
Booz Allen Hamilton
Lloyd_Kirsten@bah.com

## Abstract

As Artificial Intelligence (AI) technologies proliferate, concern has centered around the long-term dangers of job loss or threats of machines causing harm to humans. All of this concern, however, detracts from the more pertinent and already existing threats posed by AI today: its ability to amplify bias found in training datasets, and swiftly impact marginalized populations at scale. Government and public sector institutions have a responsibility to citizens to establish a dialogue with technology developers and release thoughtful policy around data standards to ensure diverse representation in datasets to prevent bias amplification and ensure that AI systems are built with inclusion in mind.

## Introduction

Data is widely-recognized as the "new oil," or the most valuable asset to organizations developing and deploying AI. While big technology companies readily share their AI methods, and often the code that goes along with them, companies are far more reticent to open up their vaults of data. As the value of data has increased, organizations have on occasion flouted the rules of acceptable practice—and even the law—in their efforts to acquire more data for their AI systems. This is further compounded by the fact that one of the greatest challenges to ensuring that AI aligns to our values is the fact that machines learn differently from humans. Because algorithms arrive at decisions using massive volumes of data, it can be nearly-impossible to parse how they make their inferences. This is particularly true of systems that rely on deep learning, which has reduced interpretability compared to other approaches.

Bias refers to undue prejudice, and in machine learning, refers to statistics that lead to a skew that inflicts an unjust outcome upon a population. Kate Crawford, co-founder and co-director of the AI Now Research Institute has discussed the mathematically specific definition of "bias" in machine learning, referring to errors in estimation or over or under representing populations in sampling. In layman's terms, biased data causes machine learning to rely on unjustified bias to discriminate against groups at scale (Crawford, 2017).

AI technologies already in the market are displaying both intentional and unintentional biases. For example, talent search technology that groups candidate resumes by demographic characteristics, or insensitive auto-fill search algorithms (Lapowsky, 2018). Datasets on their own are neither good nor bad, however, the problem lies in the ability of this bias to scale and negatively impact peoples' lives. We are now at a tipping point where if this bias is left unrecognized and unchecked, it will result in serious negative consequences that impact populations at scale. The likely ensuing backlash could stymie the progress of AI technology.

The public sector is uniquely positioned to mitigate this issue, by taking proactive steps to ensure AI systems are built with inclusion in mind. First, the government should establish comprehensive standards around data use to ensure data security and protect individual citizens' rights to privacy. Additionally, the government should release guidance around data sharing between public and private organizations to ensure data integrity by allowing for the curation of more diverse datasets, while still adhering to the aforementioned data use and security standards. Finally, the government should hold organizations accountable for adverse impacts inflicted by AI systems that amplify bias at scale through the enactment of comprehensive regulation and policy.

## Types of Bias in Machine Learning

Bias in AI / machine learning can be the result of many careless mistakes and oversights in data aggregation. Often it comes from training data sample sets that are non-representative of the general population; this means that the datasets either exclude certain groups or characteristics, or conversely, over-represent other groups. Other times it arises from human error in labeling data. In their own right, these mistakes may not seem to be overtly dangerous, however, the real issue lies in the ability for these biases or discrepancies to be magnified and inflict damage to populations on a large scale.

Bias can be broken out into five major categories- dataset bias, association bias, automation bias, interaction bias, and confirmation bias (Chou et al., 2017). Dataset bias refers to bias that results from a too-small or too homogenous dataset, which then perpetuates inaccurate generalizations. Nikon's facial recognition software demonstrated dataset bias when showed pictures of Asian people and suggested that they were blinking, which shows that the algorithm was trained primarily on faces of other ethnicities (Lee, 2009). Association bias is similar in that the data used to train an algorithm reinforces and multiplies cultural bias when applied to a larger problem. The use of Google's algorithms in online advertising was criticized when a study by Carnegie Mellon revealed that women were far less likely than men to receive ads for high-paying jobs with salaries greater than $200,000 (Spice, 2015).

Moving into the realm where there are discrepancies between human goals and the machine's understanding of the desired end result, automation bias describes situations in which the AI fails to take social or cultural factors into consideration. By reinforcing European notions of beauty, the Beauty.AI pageant showed automation bias in action, when machines overwhelmingly chose winners with light skin rather than any of the large number of dark-skinned applicants (Levin, 2016). Similarly, interaction bias occurs when machines are empowered to learn without safeguards in place to identify and exclude harmful or pernicious beliefs; Microsoft's racist and anti-Semitic chatbot Tay that had to be shut down after 24 hours of humans teaching it racial slurs is a good example of this (Vincent, 2016). Finally, confirmation bias occurs when information is oversimplified or personified and makes improper generalizations or assumptions about a group or individual. Shopping recommendations based on past purchase history often demonstrate confirmation bias by showing customers similar products (Chou et al., 2017).

## The Problem with Bias in Data

The speed at which these decisions can impact large populations, and essentially perpetrate and reinforce bias at an industrial scale, is hugely problematic. In many cases, very little thought is being given up-front to how data are collected and used, and because of the newness of the technology, there is still confusion around how machines reach their decisions. The removal of humans from certain decision-making loops means that accountability for inaccurate or offensive behavior is diluted. If a human employee, for example, treats one or a few customers disrespectfully, his/her behavior can be quickly contained with minimal harm. If, however, an AI system is used for an inappropriate purpose that offends customers, it can affect thousands or millions of people before it is contained. In some instances, this isn't a problem. For example, autonomous cyber security systems can recognize anomalous patterns in network activity inscrutable to humans—and react in real-time. If they are wrong, the only cost is their own needless attempt to block an intrusion and the annoyance of a few users. In other cases, however, AI's lack of transparency is a serious threat to individuals' livelihood and well-being, all of which pose serious risks to organizations considering applying AI. Because of this, depending on the use case or scenario, the cost of failure could be highly asymmetric to the application, which should be assessed as part of normal risk management activities.

This is most evident in applications of AI in the criminal justice system, where AI systems are being used to make determinations concerning individuals' likelihood of recidivism, and terms in jail. Many AI algorithms used for these purposes cannot be directly parsed by humans, which often aren't subject to rules which would require their developers explain their reasoning to the individuals affected by their decisions. There is an ongoing debate in technical papers arguing whether or not bias, by what specific measure, exists in these algorithms (Angwin et al. 2016). Without transparency, the technical community cannot achieve consensus on whether bias is occurring. Due to this fact, as soon as summer 2018, the European Union may legally require companies to disclose to customers how automated systems reach their decisions to ensure that there is clarity in the decision-making process (Goodman and Flaxman, 2016). The EU's transparency mitigation solution illuminates the role that the government should have in holding organizations accountable for ensuring transparency and fairness are built into AI systems.

## Impact

Arguably one of the biggest risks in applying AI is the potential for adversely impact already-marginalized groups. In spite of attempts to remove bias in datasets, it often seeps into AI systems in ways that perpetuate and exacerbate inequality among protected classes. Careful attention to infusing principles of equality into data aggregation and AI development is needed to ensure that we create fair and equitable systems. This applies to both the way that we collect and use data, and how the types of data we use can impact the system we build. All groups should be considered and represented in accordance with their specific differences to ensure equal treatment. For example, an algorithm used to determine leave policies at work could bar women from maternity leave because men cannot become pregnant. Policymakers should dedicate resources to researching and understanding how policy can be used to guide and shape AI systems to prevent such impacts from occurring (Schlabs, 2017).

The harm or effects caused by bias in data in AI applications can be classified into two types: representative harm, and allocative harm (Crawford, 2017). Representative harm occurs when systems reinforce subordination of a group along identity lines, whereas allocative harm withholds an economic opportunity or resource from a person or persons. Allocative harm is easier to identify, because it is immediate, easily quantifiable, and discrete. Representative harm, however, is harder to detect because its impact is realized over a longer time frame and is harder to formalize or diffuse. Representative harm is caused by stereotyping, recognition, denigration, under-representation, and ex-nomination (Crawford, 2017). In 2013, LaTanya Sweeney demonstrated how bias in Google's search recommendation engine inflicted both of these harms, after searching for "black-sounding names" that surfaced ads for criminal background checks. Not only did this demonstrate representative harm through clear racial profiling, but might also inflict allocative harm if employers discriminated against applicants because of the criminal association (Sweeney, 2013). The use of algorithms in hiring decisions has come up against the Americans with Disabilities Act for effectively preventing people with disabilities from being hired (O'Neil, 2016).

Unintended bias against protected groups can be partially attributed to the fact that the developer community responsible for building and training algorithms is not adequately diverse (Litt, 2017). Algorithms optimize and amplify hidden factors in data- if the data used to train algorithms is biased, the algorithm will learn that behavior as normal. This has led to issues in healthcare, for example, where applications of AI have the potential to be transformative but could also lead to negative health outcomes. If not carefully constructed, AI systems could exacerbate health conditions that disproportionally impact subpopulations. Examples include mental health problems in the LGBTQ community, higher rates of heart disease and stroke among African Americans, and higher rates of diabetes among Hispanic Americans (Hart, 2017). Data utilized in medical trials often ignores women and elderly populations and neglects pregnant women entirely (Hart, 2017). Collectively, this means that many algorithms learn from and perpetuate treatments that are best suited to white males' but may not be the best remedy for other groups. Up to this point, some semblance of intrinsic bias in both datasets and algorithms has been accepted as inevitable or an inescapable part of reality. However, at this point, it is more of an issue of leadership than a technical matter, as policymakers and leaders can take ownership to steer the technology in the right direction (Schrage, 2018). Effectively, companies today are allowed to externalize the cost of doing better and more thorough data collection to avoid biases, and instead transfer the cost onto society.

## Solutions and the Role of the Public Sector

While we are still in the early stages of machine learning and AI adoption, too much attention is being given to longer-term threats of questionable reality, such as the rise killer robots, at the expense of addressing threats that are already here today. Government leaders and public sector officials should thus capitalize on the opportunity to create a dialogue around inclusion and the importance of eliminating bias from datasets. Not only will this save organizations from embarrassment of failures that result in negative publicity, it will also ensure the least amount of representative or allocative harm befalls people. This will ensure the development of technology with a net positive impact on society, rather than stymieing potential growth of applications such as machine learning simply because of careless attention given to bias in data.

The first line of defense against creating AI systems that inflict unfair treatment is to give more attention to how datasets are constructed before operationalizing them, which means that attention to bias cannot be an afterthought. Today, it has become all too commonplace for companies to simply develop and train algorithms on their data just for the sake of deriving insight, with no forethought around the bias that could be hidden within the data. This has resulted in too many cases of companies scrambling after the fact to uncover the source of their mistakes. Addressing bias is not as simple as removing protected attributes from the data used to train algorithms, or "scrubbing to neutral." Who gets to decide what words or attributes are removed? Additionally, do we assume that neutral is what we want (Crawford, 2017)?

Race is a classic example of this phenomenon. A model could exclude race, but including a person's address could inadvertently reveal race if there is any correlation between zip code and race. Race would unintentionally be included in the model (Morgan, 2017). Another challenge with reflecting the neutrality piece is whether or not they account for historical discrepancies or under-representation. For example, take the discrepancy between the number of male vs. female CEOs in business. This example highlights the socio-technical aspect around data and bias where many issues are social (by nature) at first, and technical second. Quick fixes to these problems will only exacerbate existing issues (Crawford, 2017).

In the face of the challenge of bias, discrimination-aware machine learning has emerged as a potential way to protect equity and prevent discrimination that could arise from using algorithms in decision-making (Žliobaite, 2017). To do this, the algorithms quantify non-discrimination regulations or policies into constraints interpretable by a machine, and then models are developed based on the codified constraints. Quantifying discrimination can help remove bias from data that has been proven to be biased, incomplete, or contain discriminatory decisions, and thus promote equity in application (Žliobaite, 2017). Although still in their early stages, such studies could be the key to translating the gap between regulations and policies developed and interpreted by humans, into terms that can be interpreted and understood by machines.

Additionally, as much of today's research and implementation has existed in a vacuum, public sector officials have a unique opportunity to aid AI researchers to ensure that bias in data does not dilute the positive effects that can be gained from using AI. Facebook's Cambridge Analytica scandal highlighted the fact that private companies should be questioned in regulating their own data use practices, and that higher mechanisms for oversight and accountability are needed to govern the use of AI systems to ensure that data is used in a way that protects citizens' rights to privacy (Schaake, 2018). The government should set forth comprehensive data standards and policy to govern business organizations' data use to ensure compliance with security and privacy standards, while also fostering improved transparency in decision-making. Another way to ensure data integrity and remove bias in datasets lies in the US government opening up access to its own datasets and remove barriers for data sharing between public and private sector by developing standards for data sharing between firms; the government should also establish standards and taxonomies to ensure that datasets are machine readable and interoperable (Carter et al., 2018).

To-date, business organizations have operated without consequence to develop and implement AI solutions that afford them efficiencies and competitive advantages; this approach is no longer sustainable because of the repeated negative impacts resulting from biased data. It is incumbent upon government leaders to close the gap between AI policy and technology implementations to ensure that the benefits of AI are not withheld from marginalized groups, and that systems are built with fairness and inclusion in mind. As the technology is still new, it is too early to predict the long-term impacts of failing to consider diversity and appropriate representation in datasets, however, as AI technologies become more commonplace, biased data will only ensure that access to the benefits of the technology remains restricted for peripheral groups. It is crucial that the public sector hold the private sector accountable for AI systems development to prevent bias amplification.

# References

Angwin J.; Larson, J.; Mattu, S.; and Kirchner, L. "Machine Bias: There's software used across the country to predict future criminals. And it's biased against blacks." *ProPublica.com.* ProPublica, 23 May 2016. Web.

Carter, W.; Kinnucan, E.; Elliot, J.; Crumpler, W.; and Lloyd, K. "A National Machine Intelligence Strategy for the United States." *CSIS.org.* Center for Strategic and International Studies, 1 Mar. 2018. Web.

Chou, J.; Murillo, O.; and Ibars, R. "What the Kids' Game 'Telephone' Taught Microsoft About Biased AI." *Fast Company.com.* Fast Company, 12 Oct. 2017. Web.

Goodman, B.; and Flaxman, S. "European Union regulations on algorithmic decision-making and a 'right to explanation.'" *AI Magazine,* 38.3 (2017). Web.

Hart, R. "If you're not a white male, artificial intelligence's use in healthcare could be dangerous." *Quartz.com.* Quartz, 10 Jul. 2017. Web.

O'Neil, C. "How algorithms rule our working lives." *The Guardian.com.* The Guardian, 1 Sept. 2016. Web.

Lapowsky, I. "Google Autocomplete Still Makes Vile Suggestions." *Wired.com.* Wired. 12 Feb. 2018. Web.

Lee, O. "Camera Misses the Mark on Racial Sensitivity." *Gizmodo.com.* Gizmodo, 15 May 2009. Web.

Levin, S. "A beauty contest was judged by AI and the robots didn't like dark skin." *The Guardian.com.* The Guardian, 8 Sept. 2016. Web.

Litt, M. "Why Robots Could Soon Be Sexist." *Fortune.com.* Fortune, 18 Oct. 2017. Web.

Morgan, L. "Your Data is Biased, Here's Why." *InformationWeek.co,* InformationWeek, 11 Oct. 2017. Web.

Schaake, M. "Algorithms have become so powerful we need a robust, Europe-wide response." *The Guardian.com.* The Guardian, 4 Apr. 2018. Web.

Schlabs, E. "Machine Learning's Implications for Fairness and Justice." *The Regulatory Review.com.* Penn Program on Regulation, 3 Oct. 2017.

Schrage, M. "Is 'Murder by Machine Learning' the New 'Death by PowerPoint'?" *HBR.org.* Harvard Business Review, 23 Jan. 2018. Web.

Spice, B. "Questioning the Fairness of Targeting Ads Online." *CMU.edu.* Carnegie Mellon University, 7 Jul. 2015. Web.

Sweeney, L. "Discrimination in Online Ad Delivery." *arXiv.org.* Cornell University Library, 28 Jan. 2013. Web.

"The Trouble with Bias- NIPS 2017 Keynote- Kate Crawford." *YouTube,* 10 Dec. 2017, https://www.youtube.com/watch?v=fMym_BKWQzk.

Vincent, J. "Twitter taught Microsoft's AI chatbot to be a racist asshole in less than a day." *TheVerge.com.* The Verge, 24 Mar. 2016. Web.

Žliobaite, I. "Measuring discrimination in algorithmic decision making." Data Mining and Knowledge Discovery, 31.4, 1060-1089 (2017). Web.